\documentclass{article}
\usepackage{ijcai24}

\usepackage[dvipsnames]{xcolor}
\usepackage{multirow}
\usepackage[flushleft]{threeparttable}
\usepackage{amssymb}
\usepackage{lipsum}
\usepackage{pifont}
\newcommand{\xmark}{\ding{55}}%

\usepackage[symbol]{footmisc}

\usepackage{times}
\usepackage{soul}
\usepackage{url}
\usepackage[hidelinks]{hyperref}
\usepackage[utf8]{inputenc}
\usepackage[small]{caption}
\usepackage{graphicx}
\usepackage{amsmath}
\usepackage{amsthm}
\usepackage{booktabs}
\usepackage{algorithm}
\usepackage{algorithmic}
\usepackage[switch]{lineno}
\usepackage{makecell}
\usepackage{color}

\title{
RealDex: Towards Human-like Grasping for Robotic Dexterous Hand
}

\author{
Yumeng Liu$^{1,2*}$
\and
Yaxun Yang$^{1*}$
\and
Youzhuo Wang$^{1*}$
\and
Xiaofei Wu$^1$
\and
Jiamin Wang$^1$
\and
Yichen Yao$^1$
\and\\
Sören Schwertfeger$^1$
\and
Sibei Yang$^1$
\and
Wenping Wang$^3$
\and
Jingyi Yu$^1$
\and
Xuming He$^1$
\and
Yuexin Ma$^{1\dag}$\\
\affiliations
$^1$ShanghaiTech University, $^2$The University of Hong Kong, $^3$Texas A\&M University\\
\emails
lym29@connect.hku.hk,
\{yangyx12022,wangyzh2023,mayuexin\}@shanghaitech.edu.cn
}

\begin{document}
\twocolumn[{
\renewcommand\twocolumn[1][]{#1}%
\maketitle
\begin{center}
    \centering
    \vspace{-7ex}
    \captionsetup{type=figure}
    \includegraphics[width=\textwidth,height=5cm]{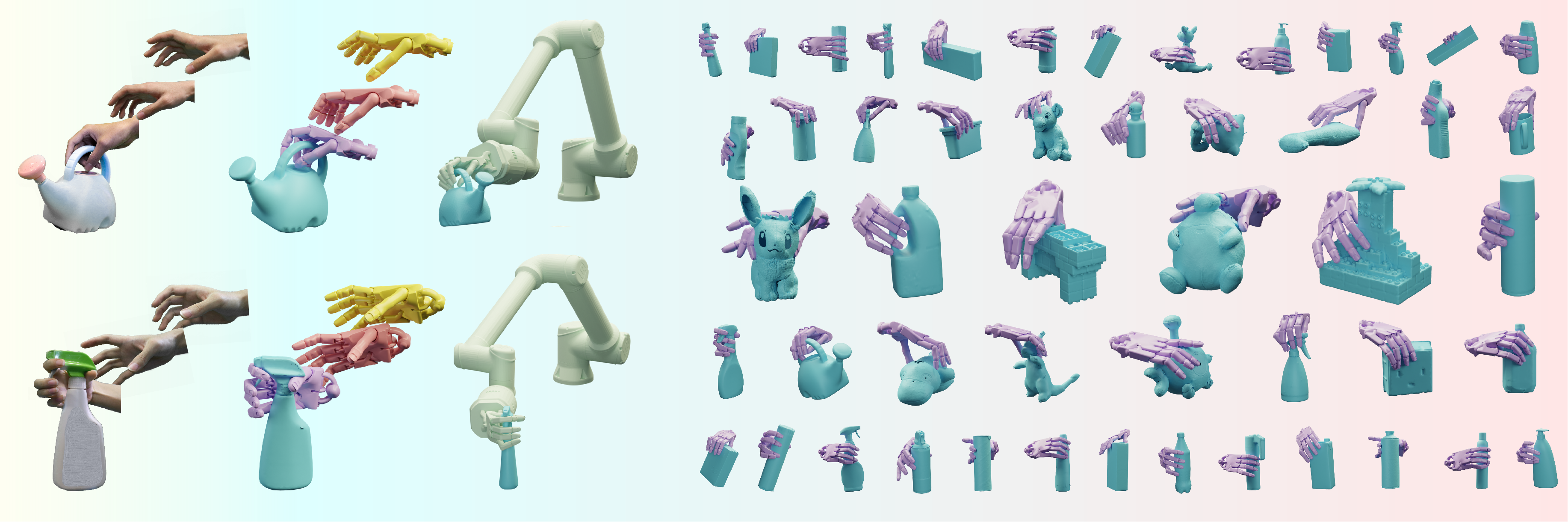}
    \vspace{-2ex}
    \caption{RealDex provides extensive realistic grasping motions of dexterous hand, synchronized with human poses and reflective of typical human motion behaviors. Utilizing the visual data of target object as input, models trained on RealDex are capable of generating human-like grasping motions for robotic hand manipulation, making them highly applicable in real-world scenarios. The right figure provides a visual gallery showcasing various objects alongside grasping poses in RealDex.}
    \label{fig:teaser}
\end{center}
}]
\begin{abstract}

In this paper, we introduce \textit{RealDex}, a pioneering dataset capturing authentic dexterous hand grasping motions infused with human behavioral patterns, enriched by multi-view and multimodal visual data. Utilizing a teleoperation system, we seamlessly synchronize human-robot hand poses in real time. This collection of human-like motions is crucial for training dexterous hands to mimic human movements more naturally and precisely. RealDex holds immense promise in advancing humanoid robot for automated perception, cognition, and manipulation in real-world scenarios. Moreover, we introduce a cutting-edge dexterous grasping motion generation framework, which aligns with human experience and enhances real-world applicability through effectively utilizing Multimodal Large Language Models. Extensive experiments have demonstrated the superior performance of our method on \textit{RealDex} and other open datasets. \textbf{The dataset and associated code are available at} \url{https://4dvlab.github.io/RealDex_page/}.

\end{abstract}
\section{Introduction}
Compared to standard parallel grippers with 7 degrees of freedom (DoF)~\cite{fang2022transcg,fang2020graspnet,dai2023graspnerf}, five-fingered dexterous hands~\cite{ShadowRobot} boasting over 24 DoF closely mimic human hand structure and movement. This sophistication enhances their ability to adapt to human-centered environments and execute more intricate and nuanced operations, such as using complex tools, with greater ease. Furthermore, robots that mirror human appearance and behavior tend to be more quickly embraced and comprehended by people, making them particularly advantageous in service and healthcare sectors.

Dexterous grasping represents an initial step in the interaction between the dexterous hand and the real world, forming the foundation for research and application of human-like service robots. Due to the high DoF of dexterous hand, traditional motion planning-based methods~\cite{andrews2013goal,bai2014dexterous,dogar2010push} struggle to handle such complex hand joint movements. In the absence of real-world data, prevalent methodologies pivot towards reinforcement learning (RL) for training dexterous hand grasping behaviors. Considering real-world applications, some of them~\cite{chen2022learning,qin2022generalizable,wan2023unidexgrasp++} have started using raw visual data, such as image or point cloud, as inputs to directly train grasping strategies for dexterous hand. However, reward functions in RL are artificially devised and cannot encompass all desired outcomes. In particular, it is difficult to model human behavioral habits in the reward mechanism. As a result, RL-trained dexterous hand grasping behaviors are only partially physically plausible and fail to truly mimic human-object interaction habits, not fully harnessing the inherent potential of humanoid robots. For example, people usually pick up a cup by its handle instead of sticking their thumb inside it, even though both ways can lift the cup physiccally. Without modeling human motion priors, current methods face notable challenges in replicating human-like grasping motions for robotic dexterous hands.

To overcome the challenges posed by data scarcity, some works~\cite{wei2022dvgg,li_gendexgrasp_2023,wang_dexgraspnet_2023,liu2020deep} have proposed synthesized dexterous grasping pose datasets. Models trained on these datasets with the supervision of ground truth can generate improved goal poses for dexterous grasping. However, these datasets, derived from simulation environments via optimization methods, rely on artificially defined energy functions and necessitate detailed physical and geometric information about objects, which are impractical for real-world applications. Furthermore, the substantial disparity between synthesized and real data makes the trained algorithms exhibit poor real-world applicability and limited generalization abilities. Additionally, the inability of simulated data to effectively model human-like behavior results in grasping poses devoid of human behavioral insights.

The teleoperation system of ShadowHand~\cite{Teleoperation} with integrated signal synchronization and control mechanisms enables the alignment of poses between the dexterous hand and human hand in real time. Utilizing this system, \textbf{we have created a comprehensive real-world dataset, \textit{RealDex}, for dexterous hand grasping}, ensuring the actions closely mirror human grasping patterns. Moreover, we have established a multi-view vision system with multiple RGB-D cameras to support research in vision-based grasping algorithms. RealDex contains 52 objects with varied scales, shapes, and materials, 2.6k sequences of grasping motions with diverse initial positions and directions, and around 955k frames of visual data, including images and point clouds. This groundbreaking dataset, driven by human hands and gathered in real-world settings, is significant for robotic dexterous hand-object grasping. It stands apart from previous datasets in three critical ways: it replicates human hand-object interactions, enhancing the training of dexterous hands to more accurately emulate human movements, thereby advancing humanoid robotics; its precise ground truth data for real dexterous hand motions bridges the gap between model training and real application; and its multi-view, multimodal visual data pave the way for the development of diverse vision-based manipulation algorithms, fostering automated perception, cognition, and action of dexterous hands in real worlds.

Leveraging the RealDex dataset, \textbf{we have developed a novel and effective method for dexterous grasping}, which involves two key stages: grasp pose generation and motion synthesis, using only the point cloud data of objects captured by real vision sensor as input. Thanks to the dataset's rich representation of human-like dexterous hand motions, our model can innately learn human behavior patterns, producing practical grasping motions. Recognizing the robust generalization capabilities of Multimodal Large Language Models (MLLMs)~\cite{Gemini,GPT-4}, we have integrated an MLLM Selection Module. This module is adept at choosing the most natural, physically plausible, and human-like grasp pose from multiple generated options, enhancing the final motion synthesis. The inherent general knowledge within large models bolsters our method's generalization capacity, ensuring effective performance even with unseen objects and thus significantly benefiting practical applications. We have conducted extensive experiments on RealDex and other open datasets, including dexterous grasping dataset and human grasping dataset, and also have tested our method on real robot hand. Quantitative and qualitative results demonstrate that our method outperforms others obviously for generation human-like practicable dexterous grasping motions.

\section{Related Work}
\subsection{Dataset for Dexterous Hand Grasping}

The complexity of dexterous hands, while allowing for precise operations, poses challenges in annotating grasping motions. Many datasets~\cite{liu2020deep,hasson2019learning,lundell2021ddgc,goldfeder2009columbia} use the planner~\cite{miller2004graspit} to generate grasping poses. However, this method's reliance on a restricted eigengrasp space limits the diversity of the data, failing to capture the full range of a multi-fingered hand's dexterity. Some datasets try to improve the quality and diversity of the grasp poses. DVGG~\cite{wei2022dvgg} utilizes MuJoCo physics simulator~\cite{todorov2012mujoco} to synthesize data and filtering out unstable grasp poses through a hand-shaking test. DexGraspNet~\cite{wang_dexgraspnet_2023} leverages a deeply accelerated differentiable force closure estimator and synthesize stable and diverse grasp poses on a large scale. MultiDex~\cite{li_gendexgrasp_2023} synthesizes versatile dexterous grasp poses across five different robot hands. Nevertheless, existing datasets for dexterous grasping consist solely of synthesized data created through optimization, resulting in a significant disconnect with real robotic hand data and the absence of human-like behavior modeling. It is noted that these datasets only contain the static grasp poses and without any dynamic grasping motion sequences. Although many datasets~\cite{yang2022oakink,chao2021dexycb,taheri2020grab} focusing on human-object interaction have emerged recently, the discrepancy between the representations of robot hand and human hand make these datasets difficult to be directly used for dexterous hand manipulation.

\subsection{Dexterous Grasping Generation}
Traditional dexterous grasping~\cite{andrews2013goal,bai2014dexterous,dogar2010push} usually leverage analytical methods to model the kinematics and dynamics of both hands and objects, performing limited for high-DoF robot hands and complex objects.
Current mainstream method for dexterous grasping is utilizing reinforcement learning (RL) for training dexterous hand grasping behaviors. Many algorithms~\cite{andrychowicz2020learning,christen2022d,huang2021generalization,she2022learning,wu2023learning} in this sphere demand exact object geometric and pose data, markedly curtailing their practical applicability. To better apply for real-world scenarios, some methods~\cite{chen2022learning,qin2022generalizable,wan2023unidexgrasp++} have begun leveraging raw visual data, like images or point clouds, to train dexterous hand grasping motions. Yet, the reward functions in RL are artificially designed and fail to model human motion patterns, making RL-based method challenging to achieve human-like grasping motions. Because our dataset provides accurate ground truth of human-like dexterous grasping motions, inspired by human motion generation methods~\cite{taheri2022goal,wu2022saga}, it is more appropriate to utilize supervised methods to guide the learning process.

\section{RealDex Dataset}
To advance research and practical applications of human-like robotic dexterous grasping in real-world environments, we introduce RealDex, the first extensive dataset that captures real dexterous hand grasping motions incorporating human behavioral patterns. RealDex features a diverse collection of 52 objects varying in scale, shape, and material, along with 2.6k sequences of grasping motions involving different initial object positions and grasping directions, and approximately 955k frames of raw visual data, including images and point clouds. We use the commercial 3D scanner, the EinScan-Pro+~\cite{EinScan-Pro+}, to reconstruct detailed 3D mesh models for collected objects. We have divided our dataset into training, validation, and test sets, ensuring that each object appears exclusively in one of these subsets. The three sets contain 2114 grasping motion sequences for 40 objects, 245 grasping motions for 6 objects, and 271 grasping motions for 6 objects, respectively. Detailed information about the division will be provided in the supplementary materials.

We will delve into the specifics of RealDex, covering aspects such as the hardware setup of the capture system in Section~\ref{subsec:system_setup}, the calibration and synchronization of all sensors in Section~\ref{subsec:calibration}, and the characteristics of RealDex in Section~\ref{subsec:Character}.


\subsection{Hardware System Setup}
\label{subsec:system_setup}

To fulfill the requirements of high-quality data collection and effective real-world application deployment, we have established a sophisticated dexterous system comprising three key components: a vision capture system designed to acquire multi-view and multimodal visual data, a manipulation system equipped with a robotic arm and a five-fingered robtic hand, and a teleoperation system that facilitates the control of the dexterous hand to emulate human-like movements. Them communicate via Robot Operation System (ROS) network.

\begin{figure}[hpt]
    \centering
    \includegraphics[width=\linewidth]{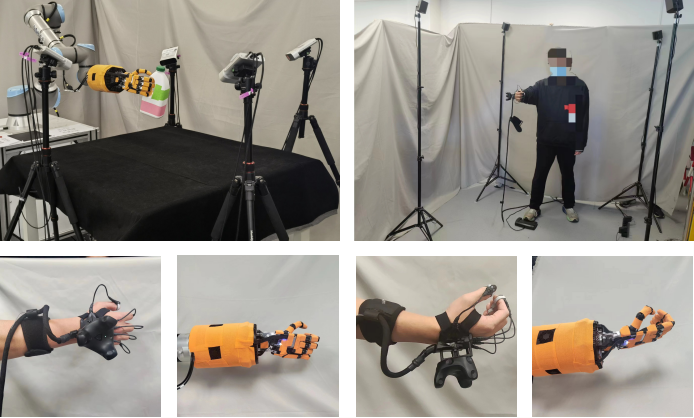}
    \caption{Overview of our comprehensive dexterous system. The bottom row shows samples of synchronized hand and robot poses.}
    \label{fig:camera}
\end{figure}

\paragraph{Vision Capture System.}
Much like human eyes, vision sensors are crucial for robots to perceive and understand their surroundings, thereby enabling appropriate action. Image-based and point cloud-based deep learning methods~\cite{qi2017pointnet,qi2017pointnet++,xu2023human,zhu2021cylindrical,wang2021fcos3d} have achieved significant breakthroughs in scene understanding, propelling the advancement of robotic perception systems. To support the practical deployment of dexterous hands, we have established a vision system equipped with four Azure Kinect RGB-D cameras with 15Hz capture frequency. These cameras are strategically positioned at four different locations, offering a quartet of perspectives focused on the operating desk, as illustrated in Figure~\ref{fig:camera}. This setup provides RGB images and point clouds from multiple angles, not only aiding in delivering precise object pose annotations for our datasets, but also enabling research and practical applications of vision-based dexterous grasping under various visual configurations, including monocular-based, multi-view-based, point-based, multimodal-based approaches, etc.

\paragraph{Dexterous Manipulation System.}
The dexterous manipulation system is designed to execute grasping operations akin to human actions. It features a UR10e robotic arm with 6 DoF and a tendon-driven Shadow Right Hand robot~\cite{ShadowRobot}. This robot hand is equipped with 20 motors, enabling adduction and abduction movements across 24 degrees of freedom. Additionally, the dexterous hand is outfitted with over 100 sensors operating at up to 1 kHz. 

\paragraph{Teleoperation System.}
For the teleoperation system~\cite{Teleoperation}, users wear a lightweight glove to intuitively control the Shadow Hand and robotic arm, allowing them to mimic natural human movements in real time. Human hand poses are seamlessly transmitted to the robot hand through the glove, vision tracking devices, and WiFi services. This system is performed by well-trained individuals to collect data on human-like dexterous grasping motions.

\begin{table*}
\centering
\begin{center}
\resizebox{\linewidth}{!}{
\begin{tabular}{c|c|c|c|c|c|c|c|c|c}
\hline
Dataset & Human &Sim/Real & Hand Type & \multicolumn{3}{c|}{Visual Data} & \#Object  & \#Grasp Pose & \#Motion Seq.   \\ \cline{5-7} 
 &  Operation    &    &           & image & point cloud &\#view &  &\\
\hline
DVGG  & \textcolor{red}\xmark & Sim & HIT-DLR II Hand  & \textcolor{red}\xmark & \textcolor{red}\xmark & - & 300 & 1.5M & - \\
MultiDex & \textcolor{red}\xmark & Sim & Multiple Types\tnote{*} & \textcolor{red}\xmark & \textcolor{red}\xmark &-  & 58 & 436k & - \\
DexGraspNet & \textcolor{red}\xmark & Sim & Shadow Hand & \textcolor{red}\xmark & \textcolor{red}\xmark & - & 5355 & 1.32M & - \\
DDGdata & \textcolor{red}\xmark & Sim & Shadow Hand  & \textcolor{red}\xmark & \textcolor{red}\xmark & - & 565 & 6.9k & - \\
\hline
RealDex& \textcolor{green}\checkmark & Real & Shadow Hand & \textcolor{green}\checkmark & \textcolor{green}\checkmark & 4 & 52 & 59k & 2630 \\
\hline
\end{tabular}
}
\caption{Comparison with existing robotic dexterous grasping datasets. Sim represents simulated data of grasp pose. \# denotes the number of the corresponding attribute. - means no related data in the dataset. Seq. denotes sequences. Multiple Types\tnote{*} includes EZGripper, Barrett, Robotiq-3F, Allegro, and ShadowHand.}
\label{tab:dataset_comp}

\end{center}
\end{table*}


\subsection{Calibration and Synchronization}
\label{subsec:calibration}

The integral dexterous system, with diverse sensors and communication networks, is inherently complex. Precise calibration and synchronization are vital to ensure seamless coordination among all subsystems for executing fine-grained operations and capturing accurate data. We introduce the most important camera-camera and robot-camera calibration and synchronization in this section.

\paragraph{Camera-camera Calibration.} Camera intrinsics comes from manufacturer, while camera extrinsics are carefully measured through a two-stage process. Firstly, we use AprilTag~\cite{krogius2019iros} to roughly calculate  extrinsics for cameras. Then, we apply the Iterative Closet Point method (ICP)~\cite{classical_icp,stytim_k4a_calibration_2023} between two neighbouring cameras to get more accurate alignment among point clouds from different views. 

\paragraph{Camera-camera Synchronization.}
The Kinect depth camera measures depth by timing the travel of emitted infrared light to objects and back. Interference from simultaneous infrared emissions by multiple cameras leads to depth inaccuracies, which we address by staggering camera sampling times and synchronizing auxiliary camera data streams with the main camera using aligned timestamps.

\begin{figure}
    \centering
    \includegraphics[width=0.9\linewidth]{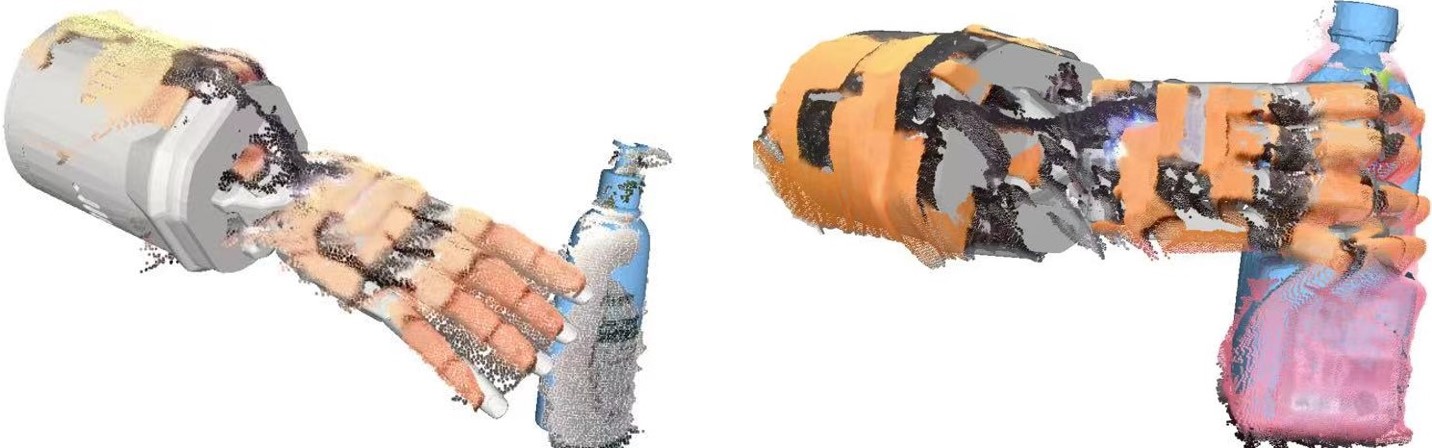}
    \caption{Visualization for well-aligned dexterous hand's mesh (white), object's mesh (blue), and the colored point cloud of the scene, demonstrating that the calibration and synchronization between robot and camera is accurate and the 6D pose estimation for object is also precise. }
    \label{fig:calibration}
\end{figure}

\paragraph{Robot-camera Calibration.}
To synchronize the robotic hand's motion with the vision system, we use hand-eye calibration~\cite{easy_handeye_2021}. This process starts with the Shadow Hand holding a calibration board, followed by moving it to over 30 locations for initial static transformation. We then refine this alignment by applying ICP method~\cite{classical_icp} between the robot's mesh vertices and the main camera's point cloud. This thorough calibration, as shown in Figure~\ref{fig:calibration}, ensures accurate alignment between the robotic and vision systems, enhancing data precision.

\paragraph{Robot-camera Synchronization.}
The robot's timestamp, derived from the ROS TF tree, continuously updates the pose of each dexterous hand joint. To ensure precise time alignment, we match the initial camera frame's point cloud with the robot's mesh, reconstructed for each timestamp within a proximate time window. We then identify the robot frame most similar to the point cloud for synchronization. With this initial aligned timestamp established, we can extrapolate and align the remaining timestamps, ensuring a consistent and accurate correlation across the entire data set. 


\label{subsec:data_process}




\subsection{Characteristics}
\label{subsec:Character}


Robotic grasping has been extensively studied for decades, serving as an essential skill for agents to engage with their surroundings and conduct manipulation tasks. Some datasets focusing on dexterous grasping, including DVGG~\cite{wei2022dvgg}, MultiDex~\cite{li_gendexgrasp_2023}, DexGraspNet~\cite{wang_dexgraspnet_2023}, and DDGdata~\cite{liu2020deep}, have already been proposed. However, our dataset, RealDex, stands apart from these existing datasets in terms of its capture process, data types, annotations, and task orientation. We draw comparisons in Table~\ref{tab:dataset_comp} and highlight three key novel features of RealDex in the following discussion. 

\paragraph{Human-like Grasping Motion.}
In the data capture process of our dataset, the dexterous hand is operated via a teleoperation system directly controlled by a human hand. This unique approach ensures that all grasping motions in RealDex authentically mirror human behavioral patterns. In contrast, previous datasets only have static grasp poses derived from physical optimization, lacking dynamic motion sequences and any real human behavioral insights. The inclusion of realistic, human-like motion data in RealDex is profoundly significant for advancing intelligent dexterous hand development, imbued with human habits. This dataset holds the potential to propel the evolution of humanoid robotics.

\paragraph{Real Robotic Dexterous Poses.}
Using a real dexterous robotic hand for data collection, our RealDex dataset accurately captures the ground truth of dexterous poses in every frame. This allows for the direct and effective application of models trained on RealDex in real-world scenarios. In contrast, previous datasets with synthesized grasp poses exhibit a substantial disparity from real-world data, leading to performance gap in translating models trained on such data into practical applications. 

\paragraph{Rich Real Visual Data.}
We have set up a vision capture system to obtain multi-view and multimodal visual data. This wealth of information is dexterous grasping for advancing research in vision-based dexterous grasping methods, which are typically more applicable and effective in real-world scenarios. The lack of genuine visual data in previous datasets significantly limits their practical utility.

\section{Methodology}
\begin{figure*}[tp]
    \centering
    \includegraphics[width=0.9\linewidth]{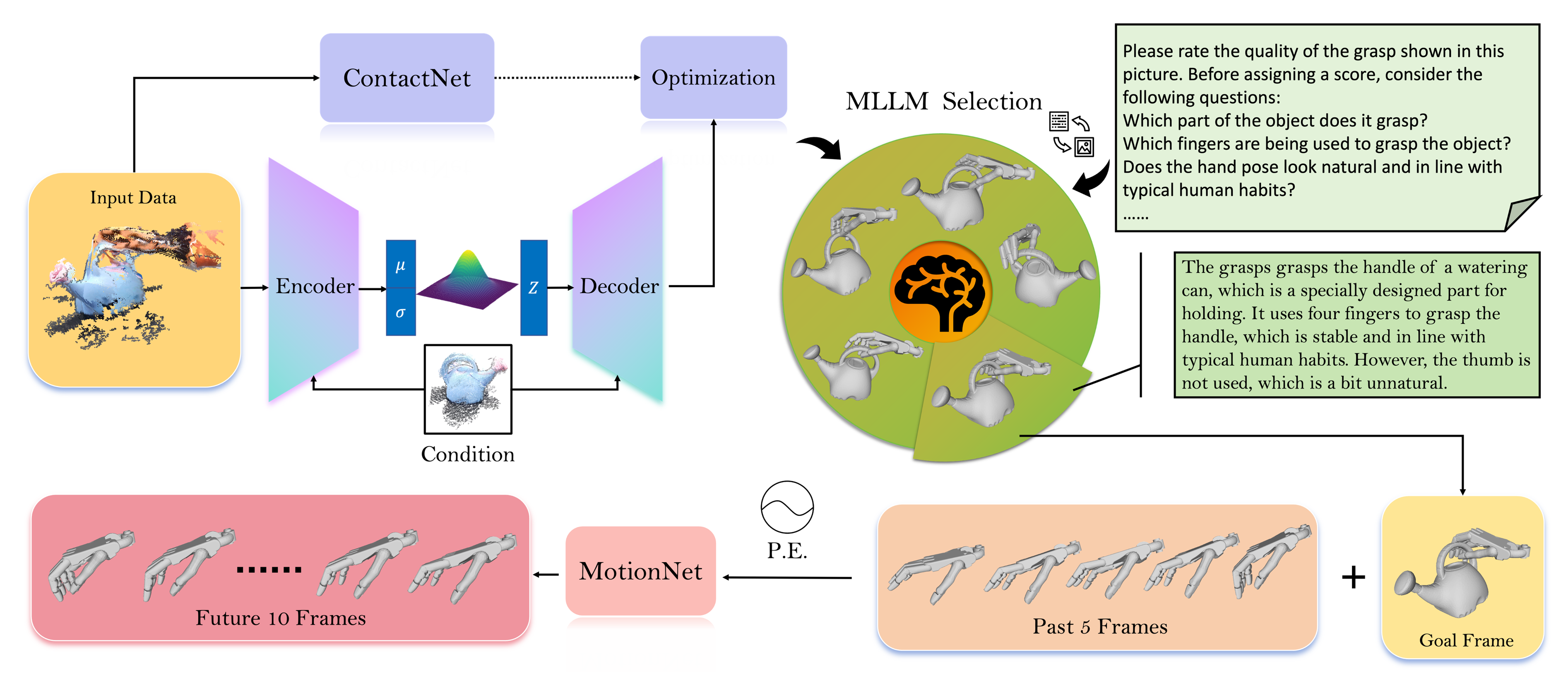}
    \vspace{-2ex}
    \caption{The architecture of our grasping motion generation framework. When observing the point cloud of the object, a cVAE-based generation module is used to generate multiple grasping pose candidates. Then, a MLLM selection module is utilized to select the most reasonable and human-like pose. Finally, based on the goal grasp pose, the MotionNet synthesizes the motion sequence for robot execution.}
    \label{fig:network}
\end{figure*}


A dexterous hand is designed to execute motions that closely resemble those performed by human hands and interactions with objects, necessitating the use of flexible strategies in different external conditions, similar to how a human would do. 
For example, the way in which humans hold a cup can vary depending on several factors, including whether there is a handle, the contents of the cup, and the temperature of the liquid inside. 
In this section, we propose a unified framework for generating complete dexterous hand motions to grasp objects, closely aligning with the human experience in various environmental settings, as depicted in Figure~\ref{fig:network}. 
Our framework primarily consists of two parts: grasping pose generation in Section~\ref{sec:method:grasp_pose} and pose-guided motion synthesis in Section~\ref{sec:method:motion_synthesis}. The former involves generating grasping poses for the input objects by using a conditional Variational Autoencoder (cVAE) to generate candidate poses and aligning them with human preferences through Multi-modal Large Language Models (MLLMs). The latter is responsible for predicting complete hand motion sequences for each pose through auto-regressive motion trajectory prediction.


\subsection{Problem Formulation}
\paragraph{Dexterous Hand Representation.} 
A dexterous hand is driven by revolute joints that only allow rotation around a fixed single axis. Hence, the pose of a dexterous hand is represented by the joint angle configuration, denoted as $\theta\in\mathbb{R}^{22}$. The global translation and orientation of the hand are represented as $\eta\in\mathbb{R}^6$, where the orientation is represented in angle-axis form. The pose of a robotic hand $\phi=(\theta, \eta) \in \mathbb{R}^{28}$ is the combination of joint angle configuration and global 6D pose of hand. The mesh representation of a dexterous hand can be generated by employing forward kinematics to calculate the global transformation for each segment. To represent the shape of hand mesh, a set of points is sampled, which we denote as $\mathbf{P}^h$. 

\paragraph{Problem Formulation.} Given an observed object point cloud $\mathbf{P}^o$, we aim to generate the preferred grasping poses $\{\phi^k\}_{k=1}^K$ and synthesize the corresponding motion sequence $\Phi = \{\phi_t^k\}_{t=1}^T$ of dexterous hand to approach and grasp object for each grasping pose $\phi^k$. 
\subsection{Grasping Pose Generation}
\label{sec:method:grasp_pose}
\paragraph{Pose Candidate Generation.}
Inspired by the work in human grasping generation~\cite{hasson_learning_2019,jiang_hand-object_2021}, we use a contact-aware generative model for dexterous grasp generation. This model uses a module to predict the contact map and a cVAE to learn a latent embedding grasping space. By sampling from the Gaussian variational distribution, the decoder of the cVAE is capable of generating a diverse array of candidate grasping poses $\{\phi^k\}_{k=1}^N$ for any input object point cloud. Subsequently, these poses undergo optimization to align with the predicted contact map, ensuring they are feasible. We then filter out unstable grasps that have insufficient contact with the object, and a human-like selection module is utilized to determine the optimal target poses $\{\phi^k\}_{k=1}^K$ for execution, where $K < N$. \\
\paragraph{Human-like Pose Selection.}
Despite the acquisition of a diverse range of candidate grasping poses through consideration of the interaction between the object and the hand, this set of candidate poses fails to account for human grasping preferences and priors. Since MLLMs~\cite{Gemini,GPT-4} have already demonstrated their capability to encode rich world knowledge, we propose using MLLMs to discern the most natural, physically plausible, and human-like grasp poses from the candidate poses, thereby aligning it with human experience. 
Specifically, we begin by acquiring mesh representations of both the robotic hand and the object for each candidate grasping pose, followed by rendering the image depicting their interaction. 
Next, we prompt Gemini~\cite{Gemini} using the specific prompt shown in Figure~\ref{fig:network}. The prompt emphasizes a comprehensive understanding of the grasping posture, highlighting the nuanced interactions between the fingers and the object. Gemini scores the rendered image, which contains intricate hand-object interaction details, in terms of naturalness, physical plausibility, human-likeness, and preference in order to select the target poses $\{\phi^k\}_{k=1}^K$.


  

\subsection{Pose-guided Hand Motion Synthesis}
\label{sec:method:motion_synthesis}

The synthesis of intermediate motion for the human hand in an auto-regressive manner, given a starting pose $\phi_0$ and a target pose $\phi_{\text{tgt}}\in \{\phi^k\}_{k=1}^K$, has been studied in previous works~\cite{zhang2021manipnet,taheri2022goal}.
Like them, we also employ an autoregressive approach, namely MotionNet, to learn how to predict future movements based on past trajectories and the current state to reach the target object. However, unlike parameterized human hand models such as MANO~\cite{MANO}, which estimate joint positions via skinning, a robotic hand's joints link its components with fixed articulatory relationships.
By strengthening the modeling of the interdependence among these joints, we can improve dexterous grasping. Therefore, we encode the joints' coordinates and utilize self-attention to model their spatial relationships. 

Specifically, for each time frame $t$, let $J_t^{\text{PE}}$ denote the sinusoidal positional encoding to the joints' coordinates. The encoded joint positions are given by
\begin{equation}
    \mathcal{F}^{J}_t = [\text{Attn}(Q,K,V)]_t,
\end{equation}
where the query $Q$ and the key-value pair $(K,V)$ are all positional code $J_t^{\text{PE}}$, and Attn means the self-attention mechanism~\cite{vaswani2017attention}. 

Next, we utilize the hand pose information from the previous five timesteps in conjunction with the current timestep's hand and object point cloud features to forecast the hand poses for the subsequent ten timesteps following~\cite{taheri2022goal}. 
At timestep $t$, the static state of a dexterous hand is represented by hand poses $\phi_t$, sampled points on the hand's mesh $\mathbf{P}_t^h$, and the joint feature $\mathcal{F}^{J}_t$. 
Joint features and hand poses from the preceding five timesteps, denoted as $\mathcal{F}^{J}_{t-5:t}$ and $\phi_{t-5:t}$ respectively, collectively characterize the motion trajectory of the hand.
The velocity of hand points at time $t$, represented by $\dot{\mathbf{P}}_t^h$ is instrumental since the velocity is directly correlated with the motion in subsequent frames. Let $\mathcal{F}^h_{\text{tgt}}$ represent the global feature of the target hand points. It is a guidance signal to encourage motion progress towards the predefined target. However, this guidance, when provided in isolation, is incomplete as it does not account for the current state of motion. To address this, we compute the displacement of the hand points from the current frame to the target frame, denoted by $\mathbf{d}^h_t$, which serves to quantify the spatial difference and provide more contextually relevant guidance toward the target. In the end, the input of MotionNet is given by:
\begin{equation}
    \mathcal{M}_{\text{in}} = (\mathcal{F}^{J}_{t-5:t}, \phi_{t-5:t},\mathbf{P}^h_t, \dot{\mathbf{P}}^h_t,  \mathcal{F}^h_{\text{tgt}},\mathbf{d}^h_t).
    \label{eq:motion_input}
\end{equation}
The MotionNet integrates a gating mechanism to encode the motion phase and utilizes MLP layers to predict the change of poses~\cite{taheri2022goal,starke2019neural}. The change can capture the temporal dynamic of movement. Representing the changes of parameters relative to current frame $t$ as $\Delta$, the output MotionNet produce is:
\begin{equation}
    \mathcal{M}_{\text{out}} = \Delta\phi_{t:t+10}, 
    \label{eq:motion_input_final}
\end{equation}
where $\mathcal{M}_{\text{out}}$ is used to reconstruct future motion and compute the input in the next timestep. 

\section{Experiments}

We developed and executed our algorithm using Python with PyTorch framework. Models were both trained and tested on an Ubuntu server, equipped with eight NVIDIA GeForce RTX 3090 GPU cards.
We leveraged Gemini~\cite{Gemini} for our MLLM selection module. \textit{\textbf{More details for the training process, inference process, loss functions are introduced in the supplementary material.}} In the following, we first show the comparision results for vision-based grasping methods, and then evaluate each stage of our method in detail. Especially, test result on real robot is also demonstrated.

\begin{table}[htbp]
\centering
\resizebox{\linewidth}{!}{
\begin{tabular}{c|ccc|ccc}
\hline
Dataset &  \multicolumn{3}{c|}{GRAB} & \multicolumn{3}{c}{RealDex} \\
\hline
Method & SAGA & GOAL & Ours & SAGA & GOAL & Ours\\
 \hline
user score ↑& 1.94 & 1.65 & 2.44 & 1.31 &2.19  &2.50 \\
 \hline
\end{tabular}
}
\vspace{-2ex}
\caption{Comparison for grasping motion generation on human grasping dataset GRAB and robotic grasping dataset RealDex. 
}
\label{tab:motion}
\end{table}
\subsection{Comparison for Dexterous Grasping}
\label{sec:comparison}
With the same point cloud data as input, we generate grasping motion sequences with different methods for the object in the test set and visualize these motions in videos.  We carry out a user study to evaluate these motions based on their humanoid characteristics, grasping stability, and hand-object interaction quality, etc. The user score for each method is derived by averaging the ratings across all samples, where rank 1 gets 3 points, rank 2 gets 2 points, and rank 3 gets 1 point. A higher score indicates better performance. We have 40 users involved. 
Due to a prior shortage of real-world robotic hand grasping data, no supervised models exist for fair comparison. Therefore, we adapt methods, SAGA~\cite{wu2021saga} and GOAL~\cite{taheri2022goal}, designed for human grasp generation to produce motions for robotic dexterous hands. In addition, to verify the generalization capability of our method, we also conduct evaluation on human hand grasping dataset GRAB~\cite{taheri2020grab}. Results in Table~\ref{tab:motion} shows that our method outperforms others obviously. 


\subsection{Evaluation for Grasping Pose Generation}

\begin{table}[htpb]
\centering
\resizebox{\linewidth}{!}{
\begin{tabular}{ccccccc}
\hline
Dataset & Method &\begin{tabular}[c]{@{}c@{}}s.i.vol. \\ ($cm^3$) ↓\end{tabular}   & \begin{tabular}[c]{@{}c@{}}p.dist. \\ ($cm$) ↓\end{tabular}  & \begin{tabular}[c]{@{}c@{}}sim.disp. \\ ($cm$) ↓\end{tabular} &\begin{tabular}[c]{@{}c@{}}user \\ score ↑\end{tabular}   \\
\hline 
\multirow{3}{*}{GRAB}  & UniDexGrasp &4.75 & 1.39 & 4.94$\pm$4.02 &1.75\\
 & ours w/o MLLM &1.95 & 1.94  & 1.02$\pm$2.10 & 1.87\\
 & ours &1.93 & 1.82  & 0.91$\pm$1.35 & 2.40\\
 \hline
\multirow{3}{*}{D.G.N.} & UniDexGrasp &5.48 & 1.17  & 4.48$\pm$4.83&1.80 \\
 & ours w/o MLLM &1.77 & 2.32 & 4.71$\pm$3.93 &1.86\\
 & ours &1.35 & 2.15  & 4.69$\pm$3.81 &2.37\\
 \hline
\multirow{3}{*}{RealDex} & UniDexGrasp &5.52 & 1.69 & 4.52$\pm$4.76 &1.75\\
 & ours w/o MLLM &1.86 & 3.53  & 4.12$\pm$4.03&2.02 \\
 & ours &1.80 & 3.34& 3.93$\pm$3.60&2.27\\
\hline
\end{tabular}
}

\caption{Comparison for grasp pose generation on three datasets. 
D.G.N is the abbreviation of DexGraspNet.}

\label{tab:pos_gen}
\end{table}
The self-intersection of robot hand and the intersection between hand and object are physically implausible, which are measured by the intersection volume (s.i.vol.) and the penetration distance (p.dist.). To measure the stability of grasping, we compute the simulation displacement (sim.disp.) following~\cite{salient_points,hasson_learning_2019}, which is defined as the average displacement of the object when the hand is stationary and the object is subjected to gravity. These metrics provide a quantitative basis for determining the realism and viability of the grasp. We generate 10 grasp poses for each object in the test set for the evaluation. In addition, we also use the user score by rendering images for randomly selected 3 grasp poses of each object for more comprehensive evaluation. 

\begin{figure}[hpt]
   \centering
   \includegraphics[width=0.7\linewidth]{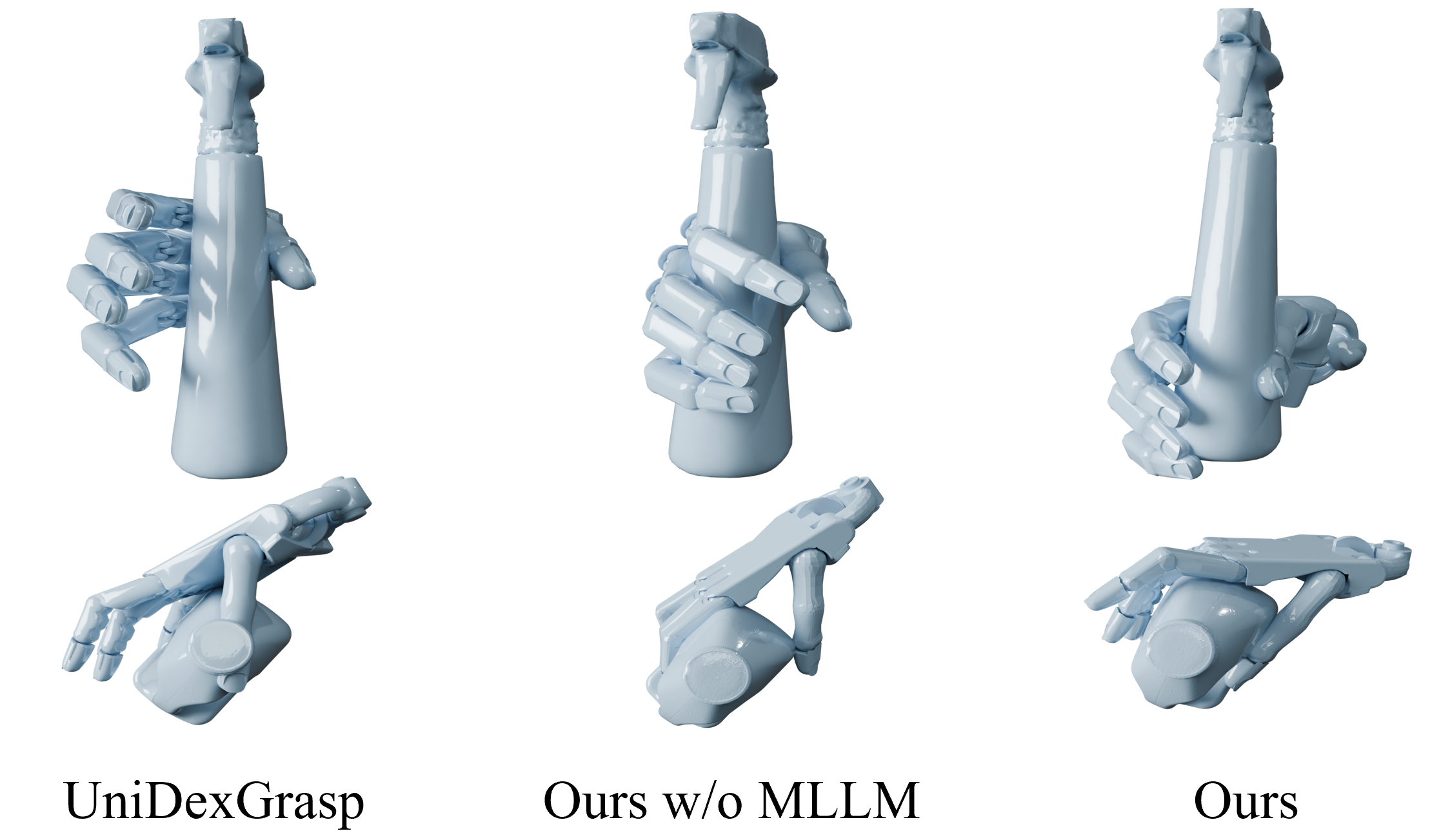}
   \vspace{-2ex}
   \caption{Visualization for grasp pose results on RealDex.}
   \label{fig:pose_abla}
 \vspace{-2ex}
\end{figure}

We compare our pose generation module with UniDexGrasp \cite{xu_unidexgrasp_2023}, the latest work for dexterous grasping. UniDexGrasp uses conditional normalizing flows to learn the distribution of plausible grasp poses. We use its released code for experiments. We evaluate on GRAB \cite{taheri2020grab}, DexGraspNet \cite{wang_dexgraspnet_2023} and our dataset RealDex. 
DexGraspNet is a synthetic dataset including various objects and grasp poses for ShadowHand. As shown in Table~\ref{tab:pos_gen}, grasp poses produced by our method demonstrate the highest stability (sim.disp.) and achieves the highest user acceptance across all datasets. The ablation result (ours w/o MLLM) demonstrates the effectiveness of our MLLM selection module in taking advantage of more general knowledge in MLLM. Qualitative result is shown in Figure~\ref{fig:pose_abla}, where our result is more approaching human-like grasp poses, more natural, and more stable.


\subsection{Evaluation for Motion Synthesis}
To purely evaluate the motion sequence quality, we input the GT goal pose and sample the same number of steps with the GT motion for evaluation. The results are shown in Table~\ref{tab:motion_syn}. Mean Per-Joint Positional Error (MPJPE) and Average Variance Error (AVE) are calculated, quantifying the mean and variance errors in predicted joint positions compared to ground truth across all frames~\cite{ghosh2021synthesis}. We also use hand mesh vertex offset and minimum distance to the object to measure the difference between final pose and the ground truth. Even with the same goal pose as the input, the motion synthesis stage in our method is still superior due to the consideration of the spatial relationship of joints.

\begin{table}[htbp]
\centering
\resizebox{\linewidth}{!}{
\begin{tabular}{cccccc}
\hline
Dataset & Method 
& \begin{tabular}[c]{@{}c@{}}MPJPE\\ ($cm$)↓ \end{tabular} 
& \begin{tabular}[c]{@{}c@{}}AVE\\($cm$)↓ \end{tabular} 
& \begin{tabular}[c]{@{}c@{}}verts.offset \\($cm$)↓ \end{tabular} 
& \begin{tabular}[c]{@{}c@{}}min.dist. \\($cm$)↓ \end{tabular} \\
\hline
\multirow{3}{*}{GRAB}  & SAGA & 1.07 & 0.1586 & 0.058  &0.057  \\& GOAL & 0.68 & 0.1261 & 0.053 & 0.025 \\
 & ours & 0.67 & 0.1035 & 0.042 & 0.022 \\

\hline
\multirow{3}{*}{RealDex} 
 & SAGA & 1.84 & 0.1902 & 0.051 & 0.048 \\
 & GOAL & 0.72 & 0.1635 & 0.049 & 0.037 \\
 & ours & 0.72 & 0.1398 & 0.046 & 0.035 \\

 \hline
\end{tabular}
}
\caption{Comparison for grasping motion generation on human grasping dataset GRAB and robotic grasping dataset RealDex. 
}
\label{tab:motion_syn}
\end{table}
\subsection{Test on Real Robot}

\begin{figure}[hpt]
   \centering
   \includegraphics[width=\linewidth]{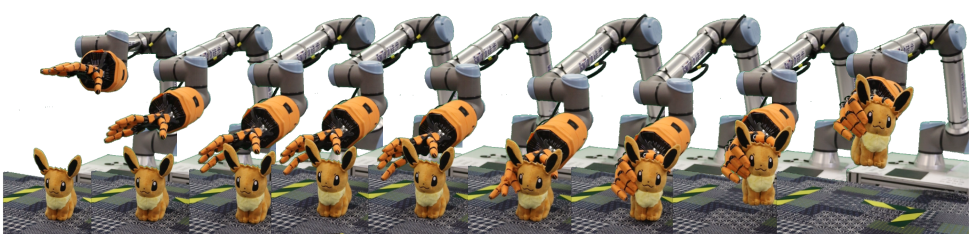}
   \caption{An example sequence of real-world grasping trajectory.}
   \label{fig:application}
\end{figure}

Because RealDex is collected by real robotic dexterous hand with accurate ground truth, our model trained on it can be directly tested on real robot, enabling fast application deployment. For motion sequences generated by our model, we send them to the simulator for safety testing before the robot execution.
The qualified motion sequence is then encoded into a stamped trajectory and seamlessly integrated into the real Shadow Hand, as shown in Figure~\ref{fig:application}. The execution process from the initial hand pose to holding the object lasts about 20s. The generated human-like grasp and accompanying approaching motion exhibit a natural appearance. Throughout the entire process, the hand and the object maintain a relatively static and stable relationship, aligning with the essential requirements for successful grasping. More videos depicting entire robot grasping process are in the appendix.


\section{Conclusion}

In this work, we present \textit{RealDex}, a groundbreaking dataset featuring genuine dexterous hand grasping motions, embedded with human behaviors and rich multi-view, multimodal visual data. Our dataset is instrumental for training dexterous hands in human-like movements, significantly enhancing humanoid robotics in real-world perception, cognition, and manipulation. We also introduce a novel dexterous grasping motion generation method using MLLM, enhancing model adaptability and real-world applicability. Extensive experiments and real robot testing show remarkable performance and practical value of our method and our dataset.

\section*{Acknowledgements}
This work was supported by NSFC (No.62206173), MoE Key Laboratory of Intelligent Perception and Human-Machine Collaboration (ShanghaiTech University), Shanghai Frontiers Science Center of Human-centered Artificial Intelligence (ShangHAI), Shanghai Engineering Research Center of Intelligent Vision and Imaging.

\section*{Contribution Statement}
* Yumeng Liu, Yaxun Yang and Youzhuo Wang contributed equally to this work. \\
\(\dagger\) Yuexin Ma supervised the project. \\
\newpage



\bibliographystyle{named}
\bibliography{ijcai24}

\newpage
\appendix
\clearpage
\setcounter{page}{1}


\section{Data Process}
\subsection{Point Cloud Denoising}
 To enhance the utility of our RealDex dataset, we have included precise mesh models and 6D poses for all featured objects. We automatically annotate the 6D poses of objects for all grasping motion sequences using the ICP method~\cite{classical_icp} and manually inspect and adjust them against the point cloud. Hardware limitations of RGBD cameras can introduce noise in the point clouds they generate, which may affect the accuracy of pose annotation. We apply a statistical outlier removal filter~\cite{Zhou2018open3d} to post-process raw point cloud data and merge point clouds from various viewpoints into a unified one. The denoising process involves analyzing each point's average distance to its 20 nearest neighbors and excluding those points whose distance deviates by more than two standard deviations from the mean, effectively reducing noise.
\subsection{Object Pose Labeling}
The 6D poses of objects are annotated mainly using the Iterative Closet Point method (ICP)~\cite{classical_icp} with human adjustment. Initially We manually determine the object pose in the first frame using the refined point cloud, setting the foundation for subsequent automated ICP adjustments. The pose for each subsequent frame is inferred from the preceding one. Finally, the resulting sequence is inspected and, if necessary, fine-tuned by a human annotator. In practice, most sequences require only a single annotation pass.

\subsection{Visualization of Dataset}
\begin{figure}[htpb]
    \centering
    \includegraphics[width=\linewidth]{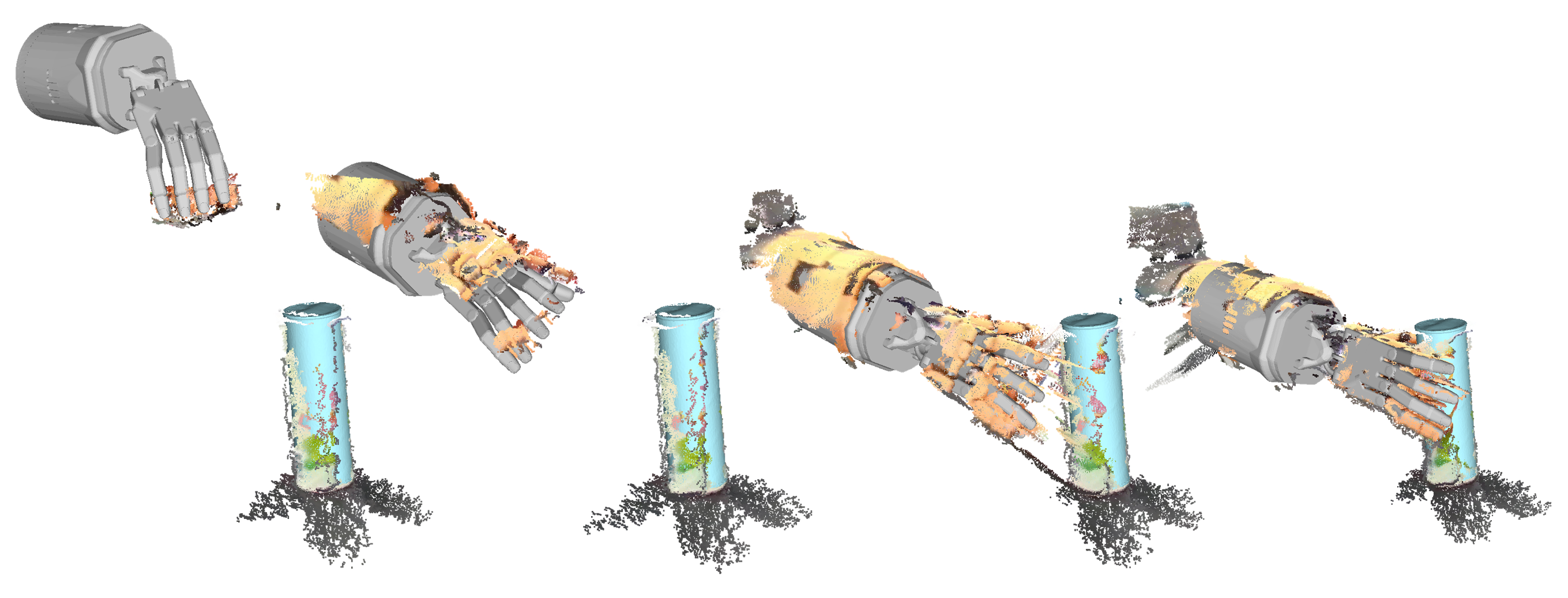}
    \caption{The visualization for aligned point cloud and hand's mesh, object's mesh.}
    \label{fig:supp_data_align}
\end{figure}

\textbf{Annotation} Here we present a sample of the annotated results depicting the object motion and dexterous hand motion, as shown
in~\autoref{fig:supp_data_align}.

\begin{figure}[htpb]
    \centering
    \includegraphics[width=\linewidth]{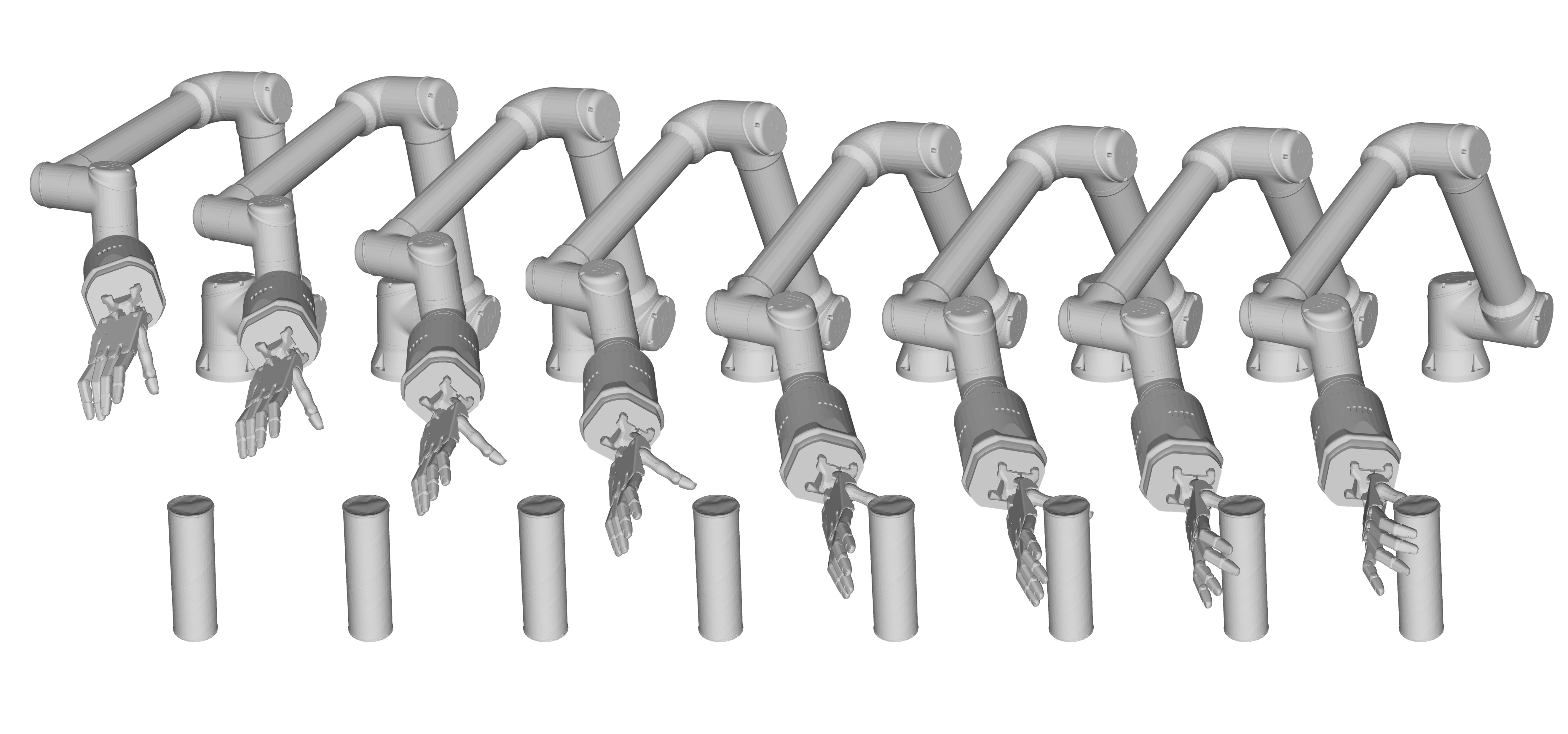}
    \caption{The visualization for grasping motion sequence in RealDex.}
    \label{fig:supp_data_motion}
\end{figure}

\textbf{Motion Sequence} We present the the motion sequence of dexterous hand mesh in our RealDex dataset. We sampled 8 frames from a grasping motion and display the mesh of robotic hand with arm, as shown in~\autoref{fig:supp_data_motion}.

\section{Method}
\subsection{Training} 
We train our framework in two stages, the training for grasp pose generation and the training for motion synthesis. Since our dataset includes precise annotations for object and hand poses along with complete dexterous hand motion, enables both stages of our training to benefit from ground truth data supervision.\\

\noindent\textbf{Pose Generation} During pose generation training, we first create the robotic hand's mesh from the hand pose $\phi$ using forward kinematics and then generate the hand's point cloud $\mathbf{P}^h$. The hand feature $\mathcal{F}^h$ and condition feature $\mathcal{F}^o$ is compressed into the latent space by cVAE encoder. Hand poses are reconstructed by the decoder using the concatenation of conditional feature and the latent code, sampled from the learned distribution. From the decoder's output, we can then compute a binary contact map, $\mathcal{C}$ on object points that indicates whether the points are within the hand's contact region. The loss to supervise the generated poses is the weighted sum of four losses:
\begin{equation}
\begin{split}
   &\mathcal{L}_{\text{KL}}=\dfrac{1}{2}(-\log{\sigma^2} - 1 + \sigma^2 +\mu^2)\\
   &\mathcal{L}_{\text{recon}} = ||\phi - \phi^{\text{gt}}||_2, \\
   &\mathcal{L}_{\text{cmap}} = \text{BCE}(\mathcal{C} - \mathcal{C}^{\text{gt}}),\\
   &\mathcal{L}_{\text{CD}} = \sum_{\mathbf{a}\in \mathbf{P}^h} \min_{\mathbf{b}\in \mathbf{P}^{h, \text{gt}}} ||\mathbf{a}-\mathbf{b}||^2 
   + 
   \sum_{\mathbf{b}\in \mathbf{P}^{h, \text{gt}}}  \min_{\mathbf{a}\in \mathbf{P}^h}||\mathbf{b}-\mathbf{a}||^2.
\end{split}   
\label{eq:loss_pose}
\end{equation}
In~\autoref{eq:loss_pose}, $\mathcal{L}_{\text{KL}}$ denotes the Kullback-Leibler divergence to measure the similarity between prior $\mathcal{N}(\mu, \sigma^2)$ and standard Gaussian distribution $\mathcal{N}(0, 1)$; $\mathcal{L}_{\text{recon}}$ is the MSE loss of reconstructed hand pose and ground truth hand pose; $\mathcal{L}_{\text{cmap}}$ is a binary cross entropy (BCE) to measure the difference between the contact map from reconstructed hand pose and the ground truth; and $\mathcal{L}_{\text{CD}}$ is the Chamfer distance between points sampled from reconstruction hand mesh and the points on GT hand mesh.

\noindent\textbf{Motion Synthesis} In the training of MotionNet, we first generate the hand points $\mathbf{P}^h$. Then we add noise to $\phi$ and $\mathbf{P}^h$ in the network input to enhance the generalization ability of network. The loss for MotionNet is the difference from predicted parameters to its GT value.
\begin{equation}
\begin{split}
   \mathcal{L}_{\text{M}} &= \omega_{\phi}||\phi - \phi^{\text{gt}}||_1 + \omega_{h}||\mathbf{P}^h - \mathbf{P}^{h,\text{gt}}||_2 + \omega_{d}||\mathbf{d}^h - \mathbf{d}^{h,\text{gt}}||_2 
\end{split} 
\end{equation}

\begin{figure*}[htpb]
    \centering
    \includegraphics[width=\linewidth]{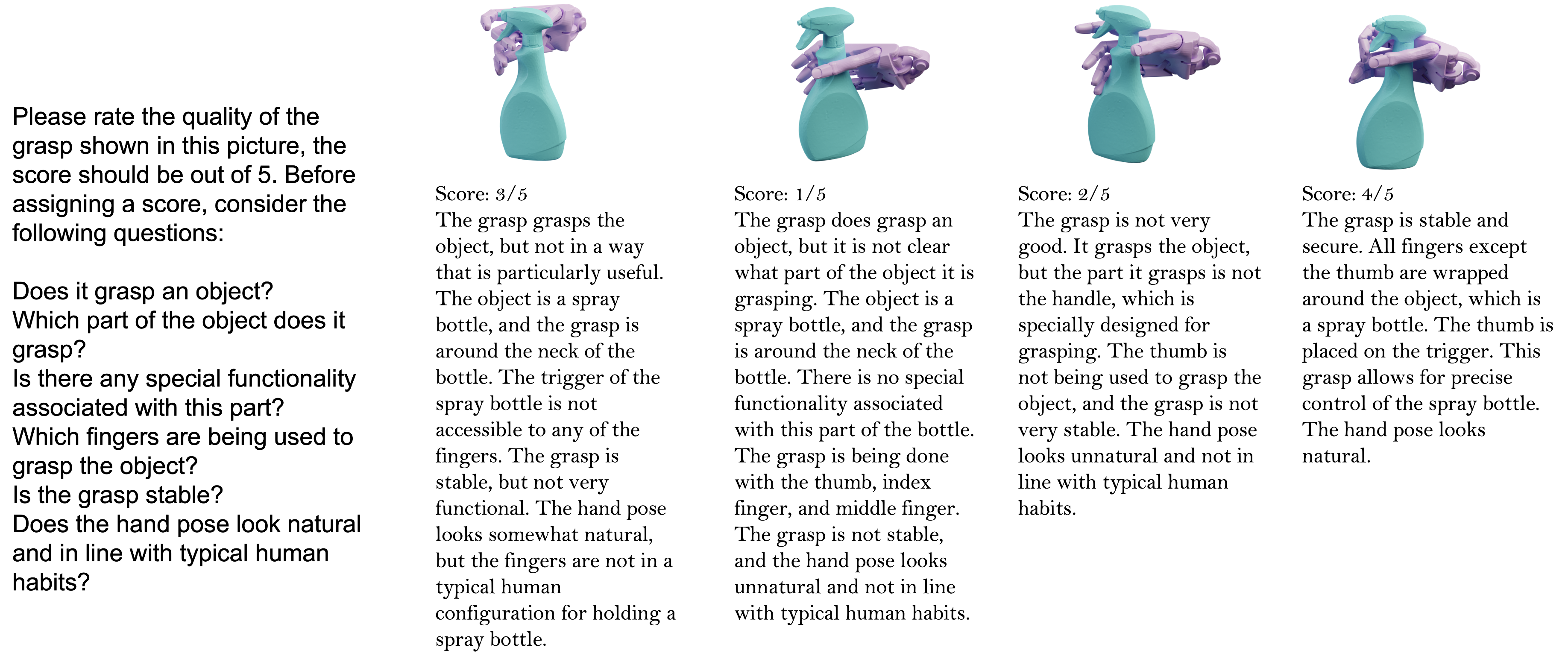}
    \caption{The text in the first column provides the complete prompt input to Gemini. Adjacent to this, in the subsequent four columns on the right, we present the input images alongside the corresponding scores and explanations as given by the MLLM selection module, offering a transparent view of the decision-making process.}
    \label{fig:supp-MLLM}
\end{figure*}

\subsection{MLLM Selection}
For each object, we sample 100 poses and generate 100 images through rendering. These images are collectively processed by  Gemini, yielding a set of scores along with detailed explanations for each pose. Subsequently, we extract the top ten poses from the dataset, which are determined by the scores they received. These selected poses serve as the primary targets for our subsequent motion synthesis phase.


\subsection{Inference} 
At the inference stage, our pose generation module receives unseen object point clouds, which serve as the input conditions. Utilizing these conditions, cVAE decoder generates candidate grasping by randomly sampling the latent code from standard Gaussian distribution. Candidate poses are refined by test-time optimization and then get scores from LLM selection module, special requirements or conditions can be added to let the LLM select the most suitable pose as goal. Finally the MotionNet utilizes the selected poses as targets and initiates the motion synthesis process from the mean pose, indicating that all joint angles of the dexterous hand are set to zero. The output for the current time frame is then employed to determine the input data for the subsequent time frame. The termination of this process is defined by either fixed time steps or a threshold based on the distance between the current grasp and the target grasp.
\begin{figure}[ht]
    \centering
    \includegraphics[width=\linewidth]{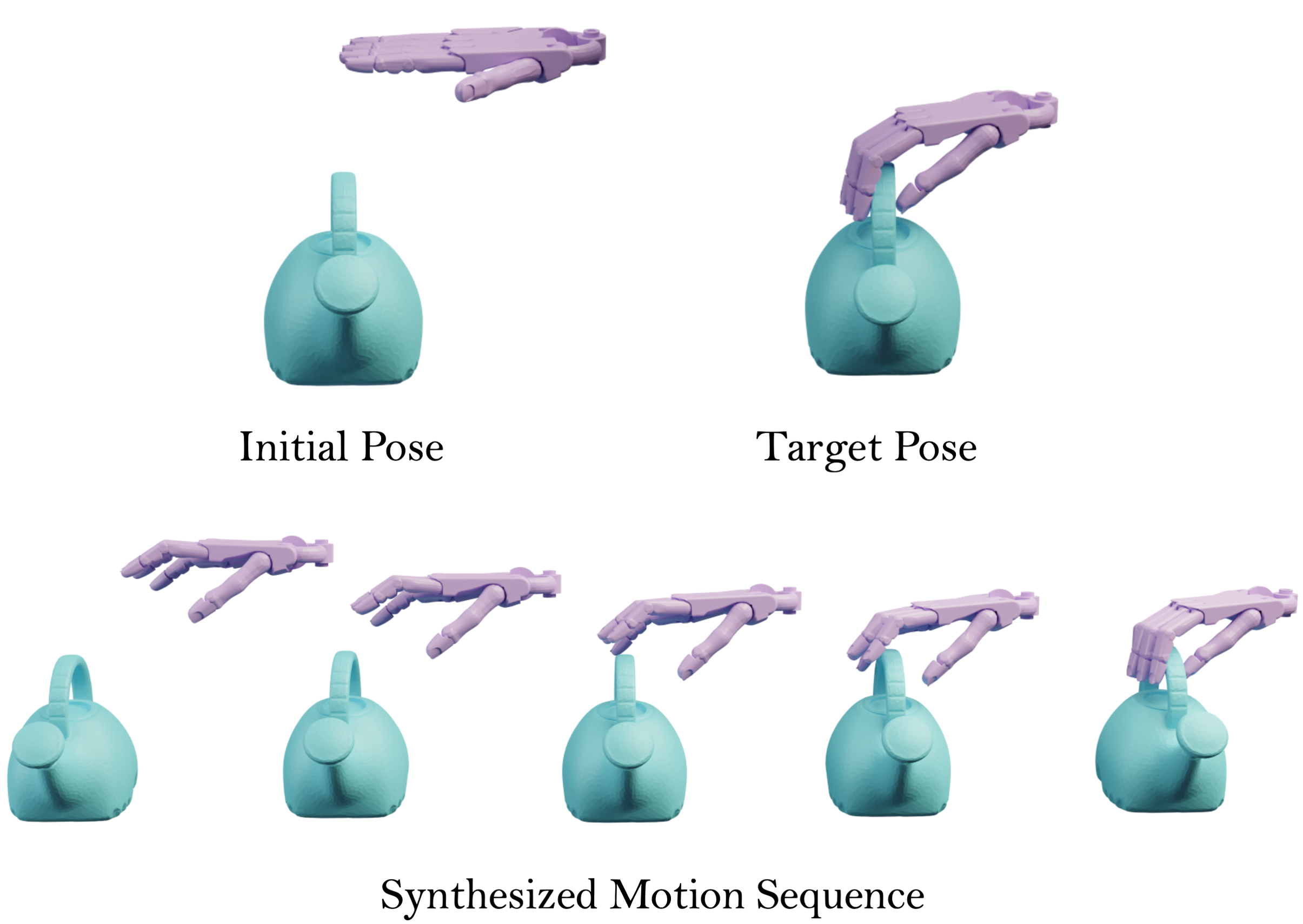}
    \caption{Motion synthesis result from our framework. The first row illustrates the initial and target hand poses, serving as inputs for the motion synthesis module. Subsequently, a sequence of hand motions is generated, using the target pose as a reference to guide the synthesis process. }
    \label{fig:supp-synth-motion}
\end{figure}
\section{More Results and Discussions}
\subsection{Pose generation}
\begin{figure*}[htpb]
    \centering
    \includegraphics[width=\linewidth]{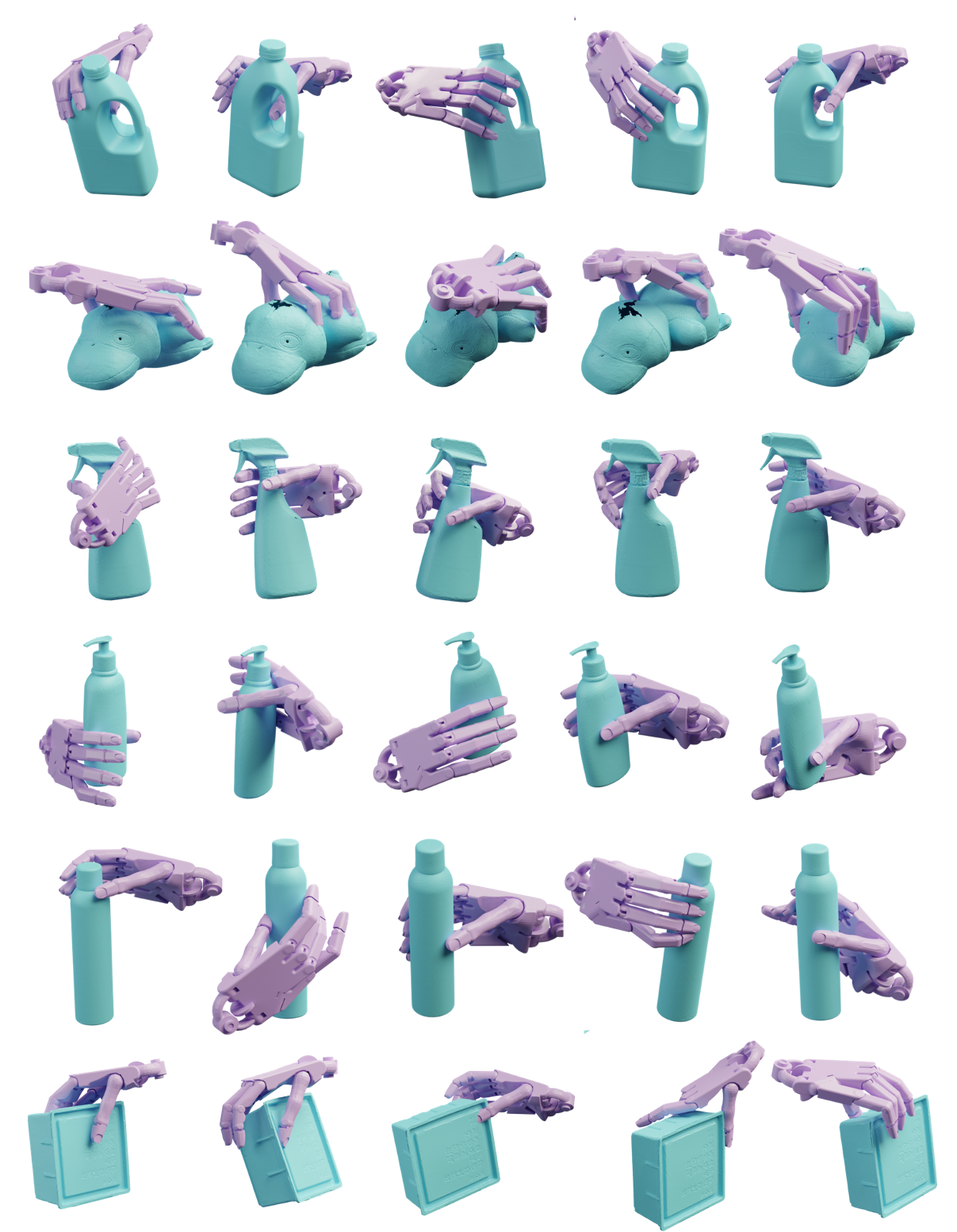}
    \caption{Visualization of the generated grasps from our grasp pose generation module. Given an object point cloud derived from RGB-D data, this module samples potential hand poses and employs MLLM to select the most plausible ones.  }
    \label{fig:supp-gallery}
\end{figure*}

\autoref{fig:supp-gallery} displays selected results from our grasping pose generation module, showcasing various automatically computed hand configurations for different object shapes.
\subsection{Motion synthesis}
Given a initial pose and a target pose, our pose-guided hand motion synthesis module is capable of generating a sequence of hand motion, as shown in~\autoref{fig:supp-synth-motion}, the initial pose we give is the mean pose of dexterous hand, which means that all the joint angles equal 0 in this pose. The translation of the hand is calculated from the average location across our dataset. Each one in the generated sequence represents a progressive step towards achieving the final target configuration.

\subsection{MLLM selection}
In~\autoref{fig:supp-MLLM}, we show the output from our MLLM selection module, each grasp is represented by a rendered image of the hand and object mesh. These images are input into the MLLM selection module, which assigns a score to each grasp and give detailed explanation.

\section{Limitation}
Our algorithm still has much room for improvement. For instance, in the result of pose generation, there is intersection between object and hand that need to be removed by optimization in test time. It could be improved by utilizing penalty loss for collision when training. In addition, when generating motion, it is guided solely by the target pose, without taking into account the actual conditions of the objects and the environment. 


\end{document}